\documentclass[letterpaper, 10 pt, conference]{ieeeconf}  

\IEEEoverridecommandlockouts                              

\usepackage{cite}
\usepackage[hidelinks]{hyperref}
\usepackage{multirow}
\usepackage{graphicx}

\title{\LARGE \bf
Cybathlon - Legged Mobile Assistance for Quadriplegics
}

\author{Carmen Scheidemann, Andrei Cramariuc, and Marco Hutter
\thanks{All authors are members of the Robotic Systems Lab, ETH Z{\"u}rich, Switzerland
        {\tt\small carmensc@ethz, crandrei@ethz.ch, mahutter@ethz.ch}}.%
}

\begin{document}

\maketitle
\thispagestyle{empty}
\pagestyle{empty}

\begin{abstract}
Assistance robots are the future for people who need daily care due to limited mobility or being wheelchair-bound. Current solutions of attaching robotic arms to motorized wheelchairs only provide limited additional mobility at the cost of increased size. We present a mouth joystick control interface, augmented with voice commands, for an independent quadrupedal assistance robot with an arm. We validate and showcase our system in the \textit{Cybathlon}~\cite{jaeger2023cybathlon} \textit{Challenges February 2024 Assistance Robot Race}, where we solve four everyday tasks in record time, winning first place. Our system remains generic and sets the basis for a platform that could help and provide independence in the everyday lives of people in wheelchairs.
\end{abstract}

\section{INTRODUCTION}

\begin{figure*}[!t]
    \centering
    \includegraphics[width=0.85\textwidth]{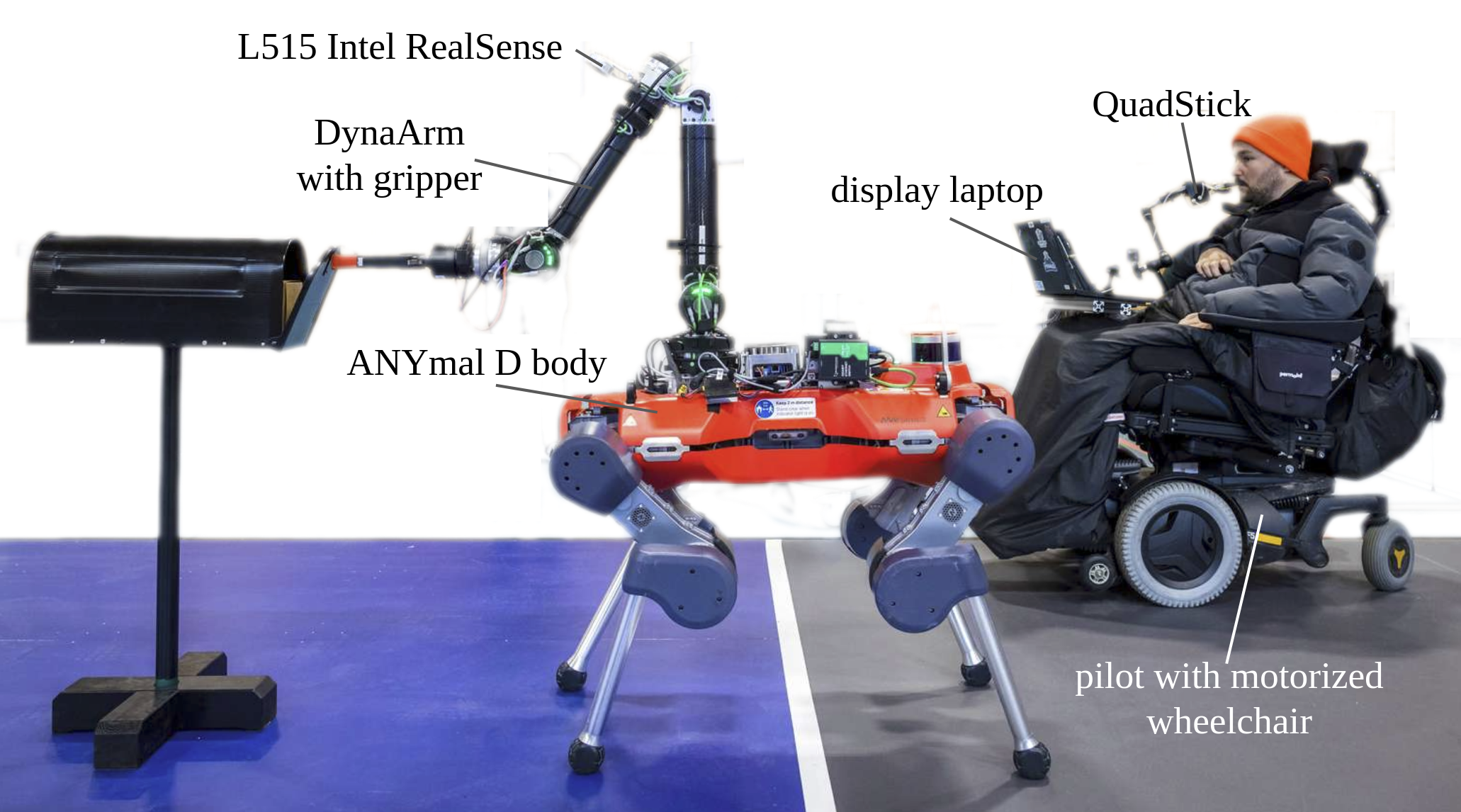}
    \caption{A overview of our proposed system in operation during the \textit{Cybathlon Challenges February 2024}. The main components of the system are an ANYmal quadrupedal base and a DynaArm robotic arm. Pictured is the pilot operating the robot to open a mailbox, which is the first task of the racetrack.}
    \label{fig:system}
\end{figure*}

People with severe motor impairments (from \textit{e.g.} muscular disease, spinal cord injury, cerebral palsy, or neurological conditions) or missing limbs struggle in day-to-day tasks. Currently, they rely on dedicated help from nurses or other caretakers. This limits their independence, which is an important aspect of health and happiness~\cite{doi:10.1080/09638280110072922}. Automated robotic solutions can provide the necessary tools for people with reduced mobility to perform more tasks on their own~\cite{doi.org/10.1155/2012/538169}, and are the future for improved quality of life and healthcare. However, these solutions are still in their early stages~\cite{doi:10.1089/rej.2017.1965}, and there is a need to both raise awareness and push towards stable platforms that can be certified.

Cybathlon~\cite{jaeger2023cybathlon, Riener2016} is a competition dedicated to showcasing and advancing technologies that help people with disabilities, \textit{e.g.}, blindness, people in wheelchairs, and prosthetic arms. The \textit{Robotic Assistance Race} (ROB) in Cybathlon focuses on a robotic assistance platform for people who are quadriplegic. The platform can take the form of either an arm attached to the wheelchair or a completely separate moving robot that the pilot can remotely control. The race consists of a set of everyday tasks that the pilot has to complete on their own, utilizing only the robotic assistant. The tasks in the February \textit{Cybathlon Challenge}~\cite{feb2024} were: collecting a package from a mailbox, brushing the pilot's teeth, hanging a scarf on a clothesline, and emptying a dishwasher.

Assistive robots will play a key role in people's lives, as current technologies are being developed for rehabilitation~\cite{7967538}, personal assistance~\cite{9223470}, and social companionship~\cite{Chapter21Companionrobotsforwellbeingareviewandrelationalframework}. Specifically for people with limited mobility who are confined to wheelchairs, most technologies focus on either smart wheelchairs~\cite{kim2023literature} or a robotic arm attached to the wheelchair~\cite{doi:10.1080/17483107.2021.2017030, 7814381, AssistiveInnovation}. With robotic arms directly attached to the wheelchair, the major limitations are reduced mobility of the equipped wheelchair and shared battery supply. In this case, the power and lifetime of the arm have to be limited to prevent the entire wheelchair from becoming inoperational. For example, the commercial iARM~\cite{10.1097, AssistiveInnovation} arm attachment from Assistive Innovations can only lift 1.5kg~\cite{AssistiveInnovation}. The size is also a limiting factor -- too big, and it becomes cumbersome to move with the wheelchair, too small, and the workspace is very limited.

Our proposed solution is the use of an independent quadrupedal robotic platform, ANYmal~\cite{7758092}, with a robotic arm, DynaArm~\cite{Dynaarm}, on top. This setup offers more versatility than mounting an arm directly on the wheelchair, as it can go places where the wheelchair can not, it has a separate power supply, and it does not limit the wheelchair's mobility or bloat its size. The independent robot can perform tasks autonomously without the pilot's direct supervision, \textit{e.g.} fetch things from another room or empty a dishwasher while the pilot does something else. It also poses less of a safety risk, as any system failures are less likely to affect the wheelchair's basic mobility, and it does not require the pilot to carry a larger battery and be in constant proximity to a moving robotic arm. To the best of our knowledge, our system is the first walking robotic personal care assistant proposed for people with limited mobility~\cite{mivseikis2020lio, bilyea2017robotic}.

We showcased our system at the \textit{Cybathlon Challenges February 2024}, where the pilot had to perform four predefined tasks in a racetrack scenario. The pilot is able to operate the robot through a mouth joystick interface, where they have direct control of its movements. Furthermore, we automate some generic actions, such as bringing an object to the mouth, which is crucial in day-to-day life \textit{e.g.,} for eating or brushing one's teeth. Finally, to enable unlimited high-level instruction without the need for tedious GUI menus, we integrate a voice command interface. Our system won in its category in the challenge, achieving a full score and a best time of $6$ minutes $34$ seconds. Although some of the task descriptions in the challenge are very specific, in this work, we present a versatile general-purpose system that can perform a variety of other tasks and actions.

\section{SYSTEM}
The system consists of three main components: (i) an ANYbotics ANYmal D body is the base of the robot, allowing for quick and robust locomotion across the track; (ii) a Dyna-Tech DynaArm, attached to the top side of the body and used for precise manipulation; (iii) a Quadstick gaming controller which is used as the main control input device by the pilot. Our system is operated by Samuel Kunz, a mechanical engineer who became quadriplegic after a swimming accident in 2014. An overview of the system and the main components is presented in Figure~\ref{fig:system}.

\subsection{ANYmal with an Arm}

The robot is a combination of an ANYbotics ANYmal D body, which serves as the quadrupedal base, and a DynaTech DynaArm~\cite{Dynaarm}, for manipulation. The ANYmal body can reach a maximum walking speed of $1.3$ m/s, allowing us to move between competition tasks quickly, and is capable of carrying a $15$ kg payload~\cite{anymalspecs}. It is largely unmodified from its original configuration, the main difference being the custom payload, which consists of an additional computer (Jetson Xavier) and the robotic arm. The DynaArm is our robotic arm of choice due to its high payload-to-weight ratio: it weighs eight kilograms and can continuously carry the same weight~\cite{Dynaarm}. It is equipped with a parallel gripper end-effector (Robotiq 2F-140) and an L515 Intel Realsense depth camera, which is mounted at the elbow joint and utilized as the primary RGB-D input for automation tasks.

\subsection{Moving the System}

We have two main ways of actuating the robot and giving it pose commands, following two different control paradigms. The first is Ocs2~\cite{9387121}, a precise MPC-based controller used for 6 Degrees of Freedom (DoF) stable control of the end-effector. Although the controller is not used for locomotion, it has access to the full system, allowing the base to tilt and translate while keeping all feet in contact with the ground. This significantly extends the robot's workspace. For example, the base can pitch down when retrieving something from the floor or lean forward to grab something far away, reducing the risk of it tipping over. Additionally, access to the full system enables it to autonomously avoid self-collision between arm and base, easing control for the pilot. Compared to learning-based full-body controllers~\cite{fu2023deep}, its model-based nature allows it to retain a stable and high-precision end-effector pose control. However, its locomotion capabilities are limited. Therefore, movement along the race track is achieved through our second control method: A learned policy, trained in simulation with reinforcement learning \cite{doi:10.1126/scirobotics.abk2822}. The policy operates in 3 DoF, accepting x, y, and yaw targets for the base. As of now, the two controllers are separate control entities which the pilot can seamlessly switch between, to either move the base or end-effector of the robot. In the future we aim to unify these two controllers into a single learned policy combining the advantages of both.

\subsection{Interfacing with the Pilot}
The pilot maneuvers himself along the race track using his own personal motorized wheelchair, to which we attach both a display laptop and a QuadStick gaming controller for the duration of the competition. The QuadStick controls will be discussed in depth in the following section \ref{sec:quadstick}. The display laptop is used to relay important information about the system to the pilot, such as the joint states of the arm, and to receive high-level input from the pilot, such as activating specific autonomy tasks with voice commands.

\subsection{Quadstick Interface}
\label{sec:quadstick}
One of our core contributions is the pilot's control interface via the Quadstick. The nature of Cybathlon as a race with extremely limited time for the completion of tasks demands a control scheme that is precise and robust while remaining intuitive for the pilot. The physical interface of the Quadstick is fairly limited and consists of a mouthpiece and a push button jointly mounted on a two-axis joystick. The mouthpiece itself contains four breath-based input channels, each of which can register no input, blowing, or sucking in a discrete manner.

A difficulty here lies in reducing the complexity of controls without compromising performance. The system possesses nine degrees of freedom, three in the base and six in the arm. These nine degrees of freedom, as well as the switching between base and end-effector controllers, need to be mapped down to the seven available input channels and the primary and secondary axis mappings of the Quadstick.

\begin{figure}[t]
    \centering
    \includegraphics[width=4.3cm]{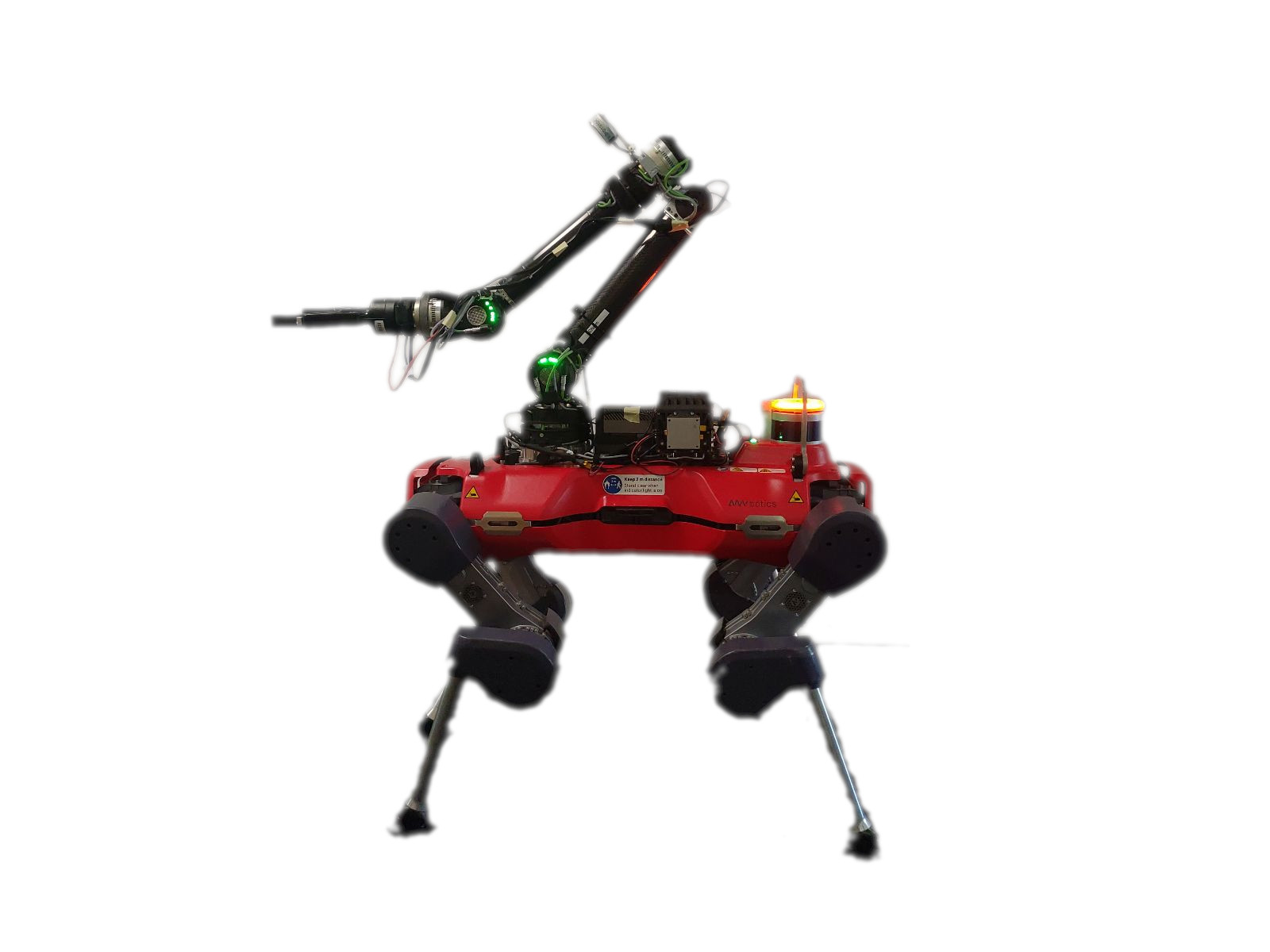}\hfill
    \includegraphics[width=4.3cm]{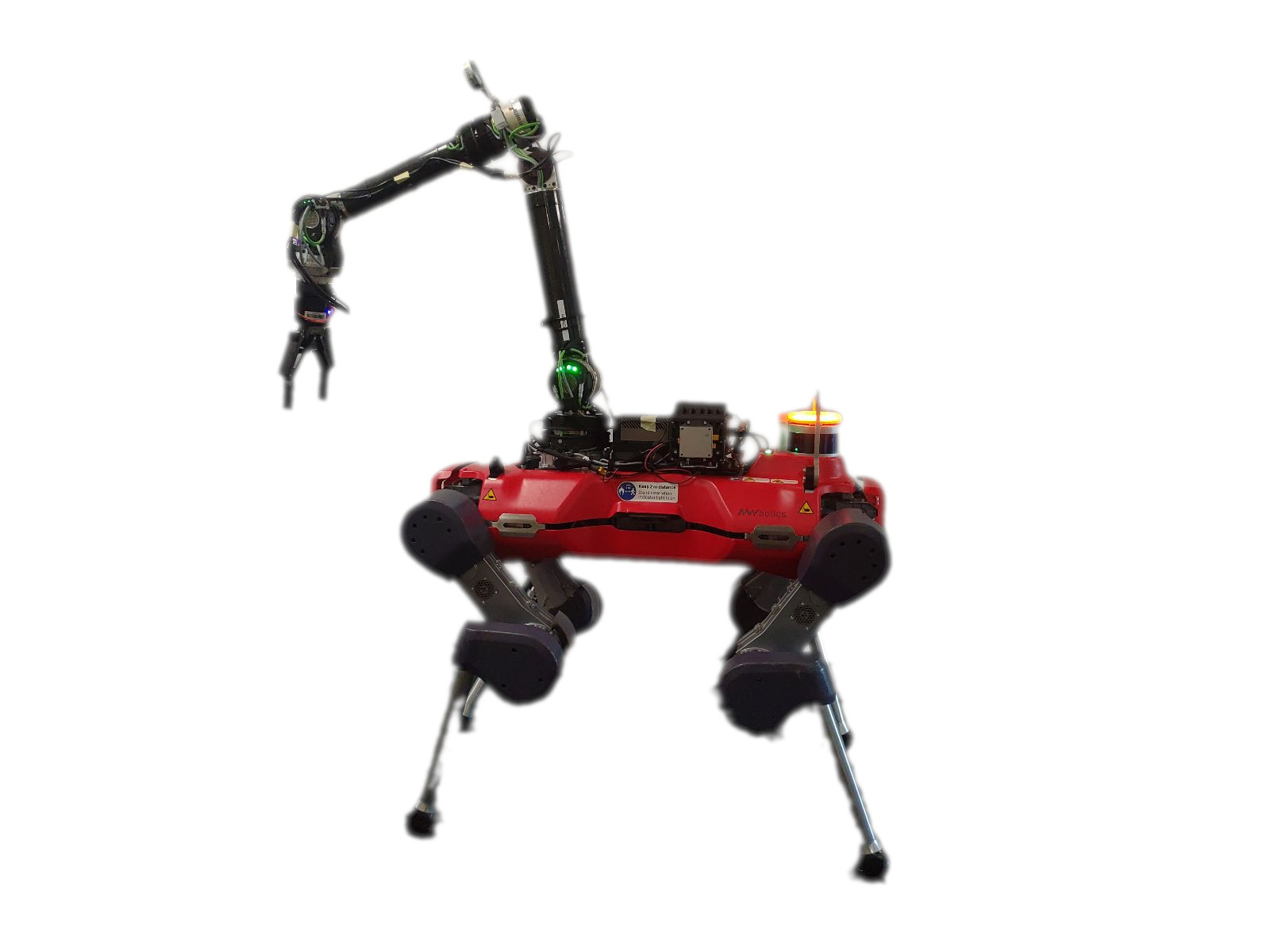}
    \caption{The initial configurations of the two control modes 
    of the Ocs2 controller. The left Figure shows that of \textit{EE Control Mode Front}, the right that of \textit{EE Control Mode Top}.}
    \label{fig:control_modes}
\end{figure}

To make the end-effector control more intuitive, we implemented two different control modes using Ocs2, which the pilot can toggle between. These differ from each other in two ways: Firstly, in their initial arm configuration, which can be seen in Figure \ref{fig:control_modes}. Secondly, they differ in their secondary joystick axis mapping, which will be elaborated upon in Section~\ref{sec:axis_mapping}. The first control mode is referred to in Table~\ref{table:axis_mapping} as \textit{EE Control Mode Front}, the second as \textit{EE Control Mode Top}. The third control mode we implement is referred to as \textit{Base Control Mode}, in which the pilot can move the base of the robot around using the learned locomotion policy while keeping the arm frozen.

The following explains the control scheme in two parts: First the layout of the Quadstick and the mode-independent input assignments are described. Subsequently we explain the mode-specific subset of channel mappings and the mapping of the primary and secondary axis assignments.

\subsubsection{Mode-Independent Inputs}

\begin{figure}[h!]
    \centering
    \includegraphics[width=8.5cm]{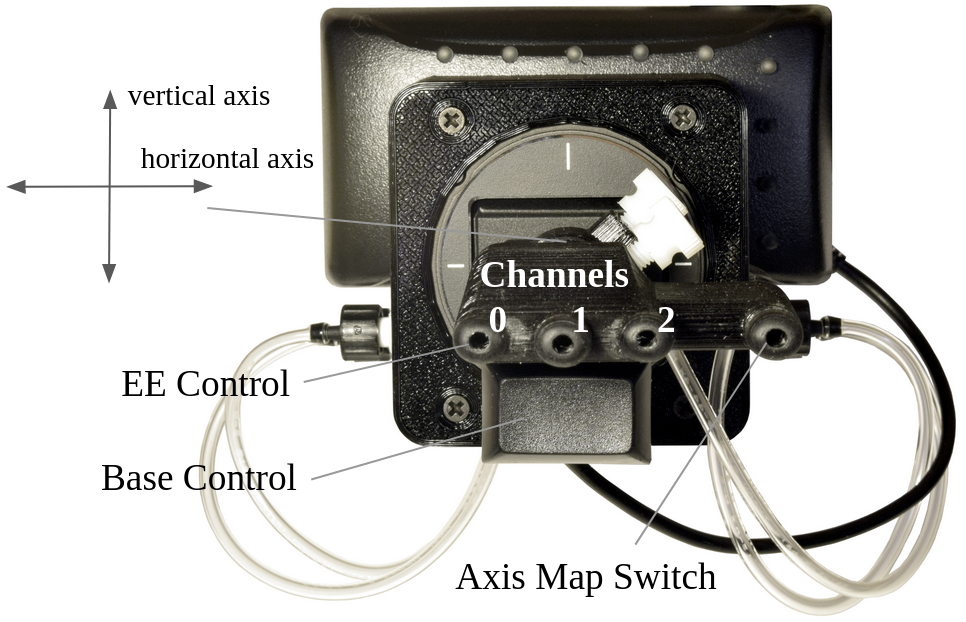}
    \caption{Quadstick Layout and Shared Input Assignments}
    \label{fig:quad_vis}
\end{figure}

The left-most breath-based input channel (referred to as Channel 0 in Figure~\ref{fig:quad_vis}) and the separate push button are used to switch between control modes. The middle-right input channel (Channel 2 in Figure~\ref{fig:quad_vis}) controls the opening and closing of the gripper. The right-most input channel (labeled \textit{Mapping Mode Switch}) can not be used directly for control as it is used internally in the Quadstick firmware to switch between the primary and secondary mapping of the joystick axis.

\subsubsection{Mode-Specific Inputs}
\label{sec:axis_mapping}

Although the Quadstick only has one physical joystick, it is possible to switch between two different axis assignments, denoted here as primary and secondary axis mapping modes, by using the rightmost breath input channel. Repeated mode switching would substantially slow down control as this switch takes a few seconds to complete. Therefore, the mappings are designed so that the majority of movement can be completed using the primary axis mapping, and the secondary is only needed for final adjustments to the manipulation angle.

The primary mapping is intuitively shared over all modes and maps the horizontal and vertical axes to x and y movement of the base or end-effector. The middle-left channel (referred to as channel 1 in Figure~\ref{fig:quad_vis}) is mapped in a controller-specific manner. In all modes, channel 1 maps the third control axis of the currently active controller, the first two being mapped by the joystick. For the base control mode, this corresponds to the yaw axis, as the x and y axes are mapped to the joystick. For end-effector control, we map this channel to the z-axis, as thus, all translational axes are exposed when in primary mapping mode.

The design of secondary axis mappings is less straightforward, and the final mapping, which can be seen in Table \ref{table:axis_mapping}, is the result of an iterative testing process with the pilot. Both the \textit{front} and \textit{top} configurations share a mapping of roll onto the horizontal axis. However, the vertical axis maps to yaw in \textit{EE Control Mode Front} and to pitch in \textit{EE Control Mode Top}. This is due to these being the most useful axes to have actuated for said configuration. For example, in \textit{EE Control Mode Front}, controlling yaw allows for grasping handles or objects that are not vertical. The secondary mode remains unmapped for the base controller.

\begin{table}[h!]
\centering
\begin{tabular}{|l|l|l|ll|}
\hline
\begin{tabular}[c]{@{}l@{}}Mapping\\ Mode\end{tabular} & \multicolumn{1}{l|}{Axis} & \multicolumn{1}{l|}{\begin{tabular}[c]{@{}l@{}}Base Control\\ Mode\end{tabular}} & \multicolumn{1}{l|}{\begin{tabular}[c]{@{}l@{}}EE Control\\ Mode Front\end{tabular}} & \begin{tabular}[c]{@{}l@{}}EE Control\\ Mode Top\end{tabular} \\ \hline
\multirow{2}{*}{Primary} & horizontal & \multicolumn{3}{c|}{x} \\ \cline{2-5} 
 & vertical & \multicolumn{3}{c|}{y} \\ \hline
\multirow{2}{*}{Secondary} & horizontal & \multicolumn{1}{c|}{-} & \multicolumn{2}{c|}{roll} \\ \cline{2-5} 
 & vertical & \multicolumn{1}{c|}{-} & \multicolumn{1}{c|}{yaw} & \multicolumn{1}{c|}{pitch} \\ \hline
\end{tabular}
\caption{Joystick Axis Mappings}
\label{table:axis_mapping}
\end{table}

\begin{figure*}[ht!]
    \centering
    \includegraphics[width=\textwidth]{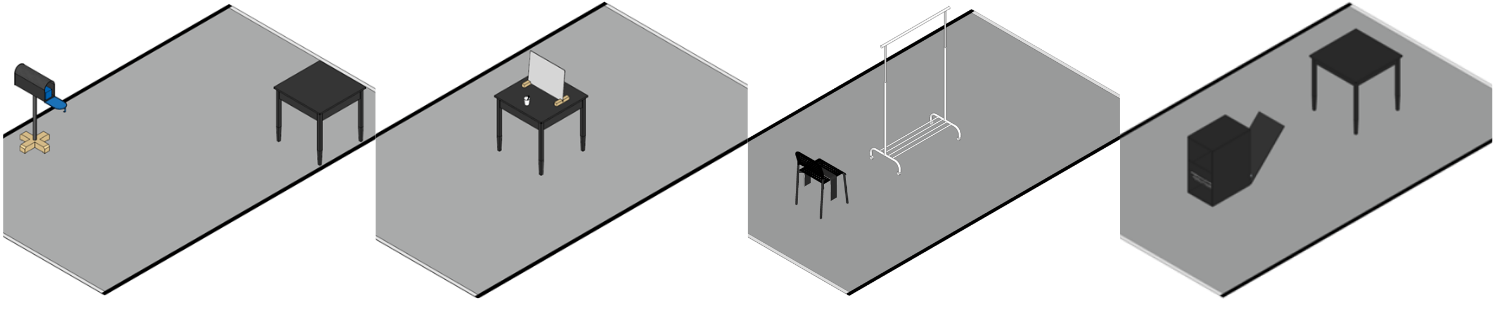}
    \caption{Visualizations of the four tasks featured in the \textit{Cybathlon 2024 February Challenge Robotic Assistance Race}. From left to right, the tasks are: (i) open a mailbox, pick the package inside, and place it on the table; (ii) pick up the toothbrush, bring it to the pilot's mouth, and place it back in the cup; (iii) pick up the scarf and hang it on the clothes rack; and (iv) open the dishwasher, retrieve a plate from inside, and place it on top.}  
    \label{fig:track}
\end{figure*}

\vspace{-15pt}

\subsection{Autonomous Face Touching}
During the \textit{February Challenge}, one of the tasks included touching the pilot's mouth with a toothbrush without the pilot moving his head toward the brush. As the Quadstick interface must be situated at the pilot's mouth, using it to teleoperate the brush into contact with the mouth is not a viable option. We therefore automate this task, keeping in mind that it is also generically applicable to other sorts of daily tasks, such as eating food.

We detect the operator's face and mouth using SCRFD~\cite{guo2021sample}. The distance to the face is calculated using the depth measurements of the L515 RGB-D camera mounted on the DynaArm. As robustness while operating in close proximity to the pilot is extremely safety-critical, additional safety measures are implemented. Firstly, the system is equipped with a force-torque sensor situated between the last link of the DynaArm and the end-effector. When in close proximity to the pilot, the trajectory-following module actively listens to potential collision events reported by the sensor, and if one is detected, it halts the approach and retracts the arm in a safe manner. Additionally, if the pilot feels the approach is unsafe, he can use voice commands to prematurely illicit a safe retraction of the arm.

\subsection{Voice Command Interface}
Using a computer interface can be difficult with limited mobility, and additional physical buttons are difficult to add and can also become confusing. To make our system more user-friendly and responsive, we use a voice command interface for activating commands that are rarely used, but necessary. For this, we first translate human speech to text using Whisper~\cite{whisper}. Then, we filter the text, searching for keywords that are mapped in a predefined manner to possible commands. For the \textit{February Challenge} the only exposed commands were: starting the face touching pipeline, interrupting face touching with a safe retreat protocol, and a general "stop" command to interrupt any current execution. For real-world deployment, using a Large Language Model (LLM) to translate natural speech into commands would provide a more intuitive solution. However, in the limited context of the Cybathlon competition, where time is a critical aspect, LLM execution is too slow.

\section{RESULTS}

As the system is mainly teleoperated by the pilot, their proficiency is integral to its performance. How much training time is needed depends largely on how intuitive the control scheme of the system is~\cite{10.1097}. During training, the pilot practiced executing the challenge tasks on a mock race track and gave feedback for the refinement of the control scheme and GUI visualization of the system. Our pilot, Samuel Kunz, trained for roughly 15 hours over five sessions to reach the level of proficiency exhibited during the \textit{February Challenge}. This included minimal failure rate executing the race tasks, collision-free movement of the robotic base across the race track, and synchronized movement of the robotic base and the motorized wheelchair.

The \textit{February Challenge} consisted of two runs, and for each team, the run with the best score (best time on equal scores between runs) was counted as the final result. The track setup is presented in Figure~\ref{fig:track}. We achieved a full score on both runs with times of 6 minutes 34 seconds and 6 minutes 51 seconds. Additionally, we were the only team to achieve a full final score. Throughout our runs, we spent on average, 23 percent of the time moving along the track and 77 percent of the time manipulating. The motion on the track includes both the robot and the pilot moving into position. While most of the time is spent on manipulation, automating many of these tasks is difficult, as making perception pipelines as reactive as direct human control is still an active challenge in the robotic community.

Although \textit{Cybathlon} is a fairly pre-defined competition scenario, our aim was to develop a generic control method that also generalizes to previously unseen tasks. The success of this can be seen in our performance on the \textit{Scarf} task, a task which the pilot had not trained for during practice on the mock course and nonetheless completed with a full score and a best time of 64 seconds.

\section{CONCLUSIONS}
We have presented the first design for controlling and operating an independent quadrupedal assistance robot for people with reduced mobility (e.g., quadriplegic). Our platform is highly versatile, as it can reach places that remain inaccessible by wheelchair and does not force the pilot to permanently carry a robotic arm attachment. This design won the \textit{Cybathlon Challenge February 2024 Assistant Robot Race}, where it completed a series of everyday tasks in record time. In the future, when assistance robots are medically certified and can perform more tasks autonomously, they will have a huge impact on the lives of many people who need daily care.

\addtolength{\textheight}{-10.5cm}   



\bibliographystyle{IEEEtran}
\bibliography{root.bib}

\begin{thebibliography}{10}
\providecommand{\url}[1]{#1}
\csname url@rmstyle\endcsname
\providecommand{\newblock}{\relax}
\providecommand{\bibinfo}[2]{#2}
\providecommand\BIBentrySTDinterwordspacing{\spaceskip=0pt\relax}
\providecommand\BIBentryALTinterwordstretchfactor{4}
\providecommand\BIBentryALTinterwordspacing{\spaceskip=\fontdimen2\font plus
\BIBentryALTinterwordstretchfactor\fontdimen3\font minus \fontdimen4\font\relax}
\providecommand\BIBforeignlanguage[2]{{%
\expandafter\ifx\csname l@#1\endcsname\relax
\typeout{** WARNING: IEEEtran.bst: No hyphenation pattern has been}%
\typeout{** loaded for the language `#1'. Using the pattern for}%
\typeout{** the default language instead.}%
\else
\language=\csname l@#1\endcsname
\fi
#2}}

\bibitem{jaeger2023cybathlon}
L.~Jaeger, R.~d.~S. Baptista, C.~Basla, P.~Capsi-Morales, Y.~K. Kim, S.~Nakajima, C.~Piazza, M.~Sommerhalder, L.~Tonin, G.~Valle, \emph{et~al.}, ``{How the CYBATHLON Competition Has Advanced Assistive Technologies},'' \emph{Annual Review of Control, Robotics, and Autonomous Systems}, vol.~6, pp. 447--476, 2023.

\bibitem{doi:10.1080/09638280110072922}
A.~Lau and K.~McKenna, ``{Perception of Quality of Life by Chinese elderly persons with stroke},'' \emph{Disability and Rehabilitation}, vol.~24, no.~4, pp. 203--218, 2002.

\bibitem{doi.org/10.1155/2012/538169}
A.~J. Pearce, B.~Adair, K.~Miller, E.~Ozanne, C.~Said, N.~Santamaria, and M.~E. Morris, ``{Robotics to Enable Older Adults to Remain Living at Home},'' \emph{Journal of Aging Research}, 2012.

\bibitem{doi:10.1089/rej.2017.1965}
L.~Penteridis, G.~D'Onofrio, D.~Sancarlo, F.~Giuliani, F.~Ricciardi, F.~Cavallo, A.~Greco, I.~Trochidis, and A.~Gkiokas, ``{Robotic and Sensor Technologies for Mobility in Older People},'' \emph{Rejuvenation Research}, vol.~20, no.~5, pp. 401--410, 2017.

\bibitem{Riener2016}
R.~Riener, ``{The Cybathlon promotes the development of assistive technology for people with physical disabilities},'' \emph{Journal of NeuroEngineering and Rehabilitation}, 2016.

\bibitem{feb2024}
\BIBentryALTinterwordspacing
{CYBATHLON Challenges 2024}. {Accessed: 2024-04-25}. [Online]. Available: \url{https://cybathlon.ethz.ch/en/events/challenges/challenges-2024}
\BIBentrySTDinterwordspacing

\bibitem{7967538}
B.~Li, G.~Li, Y.~Sun, G.~Jiang, J.~Kong, and D.~Jiang, ``A review of rehabilitation robot,'' in \emph{Youth Academic Annual Conference of Chinese Association of Automation (YAC)}, 2017, pp. 907--911.

\bibitem{9223470}
S.~Cooper, A.~Di~Fava, C.~Vivas, L.~Marchionni, and F.~Ferro, ``{ARI: the Social Assistive Robot and Companion},'' in \emph{IEEE International Conference on Robot and Human Interactive Communication (RO-MAN)}, 2020, pp. 745--751.

\bibitem{Chapter21Companionrobotsforwellbeingareviewandrelationalframework}
A.~Ruggiero, D.~Mahr, G.~Odekerken-Schröder, T.~R. Spena, and C.~Mele, \emph{Chapter 21: Companion robots for well-being: a review and relational framework}.\hskip 1em plus 0.5em minus 0.4em\relax Edward Elgar Publishing, 2022, pp. 309 -- 330.

\bibitem{kim2023literature}
Y.~Kim, B.~Velamala, Y.~Choi, Y.~Kim, H.~Kim, N.~Kulkarni, and E.-J. Lee, ``{A Literature Review on the Smart Wheelchair Systems},'' 2023.

\bibitem{doi:10.1080/17483107.2021.2017030}
J.~B. Julie~Bourassa, Julie~Faieta and F.~Routhier, ``Wheelchair-mounted robotic arms: a survey of occupational therapists’ practices and perspectives,'' \emph{Disability and Rehabilitation: Assistive Technology}, vol.~18, no.~8, pp. 1421--1430, 2023.

\bibitem{7814381}
A.~Sivaprakasam, H.~Wang, R.~A. Cooper, and A.~M. Koontz, ``{Innovation in Transfer Assist Technologies for Persons with Severe Disabilities and Their Caregivers},'' \emph{IEEE Potentials}, vol.~36, no.~1, pp. 34--41, 2017.

\bibitem{AssistiveInnovation}
\BIBentryALTinterwordspacing
{Assistive Innovations iARM}. [Online]. Available: \url{https://www.assistive-innovations.com/en/robotic-arms/iarm}
\BIBentrySTDinterwordspacing

\bibitem{10.1097}
C.-S. Chung, H.~Wang, M.~Hannan, D.~Ding, A.~Kelleher, and R.~Cooper, ``{Task-Oriented Performance Evaluation for Assistive Robotic Manipulators: A Pilot Study},'' \emph{American Journal of Physical Medicine \& Rehabilitation}, vol.~96, p.~1, 2016.

\bibitem{7758092}
M.~Hutter, C.~Gehring, D.~Jud, A.~Lauber, C.~D. Bellicoso, V.~Tsounis, J.~Hwangbo, K.~Bodie, P.~Fankhauser, M.~Bloesch, R.~Diethelm, S.~Bachmann, A.~Melzer, and M.~Hoepflinger, ``{ANYmal - a highly mobile and dynamic quadrupedal robot},'' in \emph{IEEE/RSJ International Conference on Intelligent Robots and Systems (IROS)}, 2016, pp. 38--44.

\bibitem{Dynaarm}
\BIBentryALTinterwordspacing
Dynatech: Dynaarm. [Online]. Available: \url{https://dyna-tech.ch/robotic-arm/}
\BIBentrySTDinterwordspacing

\bibitem{mivseikis2020lio}
J.~Mi{\v{s}}eikis, P.~Caroni, P.~Duchamp, A.~Gasser, R.~Marko, N.~Mi{\v{s}}eikien{\.e}, F.~Zwilling, C.~De~Castelbajac, L.~Eicher, M.~Fr{\"u}h, \emph{et~al.}, ``{Lio-A Personal Robot Assistant for Human-Robot Interaction and Care Applications},'' \emph{IEEE Robotics and Automation Letters}, vol.~5, no.~4, pp. 5339--5346, 2020.

\bibitem{bilyea2017robotic}
A.~Bilyea, N.~Seth, S.~Nesathurai, and H.~Abdullah, ``{Robotic assistants in personal care: A scoping review},'' \emph{Medical engineering \& physics}, vol.~49, pp. 1--6, 2017.

\bibitem{anymalspecs}
\BIBentryALTinterwordspacing
ANYbotics, ``{Generation D ANYmal Technical Specifications},'' May 2022, {Accessed: 2024-04-25}. [Online]. Available: \url{https://www.anybotics.com/anymal-technical-specifications.pdf}
\BIBentrySTDinterwordspacing

\bibitem{9387121}
J.-P. Sleiman, F.~Farshidian, M.~V. Minniti, and M.~Hutter, ``{A Unified MPC Framework for Whole-Body Dynamic Locomotion and Manipulation},'' \emph{IEEE Robotics and Automation Letters}, vol.~6, no.~3, pp. 4688--4695, 2021.

\bibitem{fu2023deep}
Z.~Fu, X.~Cheng, and D.~Pathak, ``{Deep whole-body control: Learning a unified policy for manipulation and locomotion},'' in \emph{Conference on Robot Learning (CoRL)}, 2023, pp. 138--149.

\bibitem{doi:10.1126/scirobotics.abk2822}
T.~Miki, J.~Lee, J.~Hwangbo, L.~Wellhausen, V.~Koltun, and M.~Hutter, ``Learning robust perceptive locomotion for quadrupedal robots in the wild,'' \emph{Science Robotics}, vol.~7, no.~62, p. eabk2822, 2022.

\bibitem{guo2021sample}
J.~Guo, J.~Deng, A.~Lattas, and S.~Zafeiriou, ``{Sample and Computation Redistribution for Efficient Face Detection},'' \emph{arXiv preprint arXiv:2105.04714}, 2021.

\bibitem{whisper}
\BIBentryALTinterwordspacing
Open-AI, ``{Introducing Whisper},'' {Accessed: 2024-04-26}. [Online]. Available: \url{https://openai.com/research/whisper}
\BIBentrySTDinterwordspacing

\end{thebibliography}


\end{document}